\documentclass[a4paper,twoside,french]{article}
\usepackage{rfiap2020}
\usepackage[T1]{fontenc}
\usepackage{babel}
\usepackage{times}
\usepackage{graphicx}
\usepackage{hyperref}
\usepackage{xcolor}
\usepackage{float}

\begin{document}
\date{}
\title{\Large\bf Apprentissage profond pour la classification de QR Codes bruités
       }
\author{\begin{tabular}[t]{c@{\extracolsep{6em}}c@{\extracolsep{6em}}c}
Rebecca Leygonie  & Sylvain Lobry & Laurent Wendling\\
\end{tabular}
{} \\
 \\
        Université de Paris, LIPADE, F-75006 Paris, France   \\
{} \\
 \\
Rebecca.leygonie@etu.u-paris.fr\\
}
\maketitle
\thispagestyle{empty}
\subsection*{R\'esum\'e}
{\em
Nous souhaitons définir les limites d'un modèle classique de classification fondé sur l'apprentissage profond lorsqu'il est appliqué sur des images abstraites, qui ne représentent pas d'objets identifiables visuellement.
Les QR Codes 
entrent dans cette catégorie d'images abstraites : un bit correspondant à un caractère encodé, les QR codes n'ont pas été conçus pour être décodés à l'oeil nu. Pour comprendre les limites d'un modèle basé sur l'apprentissage profond pour la classification d'images abstraites, nous entraînons un modèle de classification d'images sur des QR Codes générés à partir des informations obtenues lors de la lecture d'un pass sanitaire. Nous comparons les performances d'un modèle de classification avec celles d'une méthode classique de décodage (déterministe) en présence de bruit. Cette étude nous permet de conclure qu'un modèle basé sur l'apprentissage profond peut être pertinent pour la compréhension d'images abstraites.
}
\subsection*{Mots Clef}
Classification d'images, Apprentissage profond, QR Code, 
Images abstraites, Robustesse au bruit, Données de santé

\subsection*{Abstract}
{\em 
We wish to define the limits of a classical classification model based on deep learning when applied to abstract images, which do not represent visually identifiable objects.
QR codes (Quick Response codes) fall into this category of abstract images: one bit corresponding to one encoded character, QR codes were not designed to be decoded manually. To understand the limitations of a deep learning-based model for abstract image classification, we train an image classification model on QR codes generated from information obtained when reading a health pass. We compare a classification model with a classical (deterministic) decoding method in the presence of noise. This study allows us to conclude that a model based on deep learning can be relevant for the understanding of abstract images.
}
\subsection*{Keywords}
Image classification, Deep learning, QR Code, 
Abstract images, Noise robustness, Health data

\section{Introduction}
Depuis plusieurs années, les modèles du domaine de la vision par ordinateur montrent de très bonnes performances sur des images représentant des objets identifiables visuellement. De multiples architectures basées sur les réseaux de neurones convolutifs (CNN) ont vu le jour comparant leurs performances notamment sur le jeu de données ImageNet\footnote{https://www.image-net.org/}\cite{ImageNet} pour le concours ImageNet Large Scale Visual Recognition Challenge (ILSVRC)\footnote{https://www.image-net.org/challenges/LSVRC/}\cite{ILSVC} de 2010 à 2017.\\
Depuis 2012, l'utilisation de l'apprentissage profond et plus précisément des CNN s'est répandue et le taux d'erreur, mesure d'évaluation des modèles dans le concours ILSVRC, a considérablement diminué. En effet, en 2012 AlexNet \cite{AlexNet} a réduit le taux d'erreur de 16\%, suivi de ZFNet \cite{ZFNet} en 2013 (-12\%), VGG \cite{VGG} (-7,3\%) et GoogleNet \cite{Google} (-6,7\%) en 2014, ResNet \cite{resnet} en 2015 (-3,6\%), etc.\\
Les images du jeu de données ImageNet représentent des objets identifiables à l'oeil nu comme des objets du quotidien ou des animaux. Ces images sont dites figuratives.\\

Parallèlement, de nouvelles représentations de l'information sont de plus en plus utilisées au quotidien. C'est le cas notamment des QR Codes (Quick Response codes) \cite{qr} qui encodent une chaîne de caractères en bits représentés par une grille de carrés de couleur foncée ou claire. Les QR codes et ses dérivés sont largement utilisés à des fins commerciales au quotidien mais aussi pour récupérer des informations sécurisées comme avec le pass sanitaire, utilisé dans le cadre de la lutte contre la COVID-19. Les informations encodées dans ce type de représentation ne sont pas identifiables à l'oeil nu. Nous définissons ces images comme étant abstraites. Dans ce travail, nous souhaitons définir les limites d'un modèle basé sur un CNN pour la classification d'images abstraites.\\

\begin{figure*}[!ht]
    \centering
    \includegraphics[width=.9\textwidth]{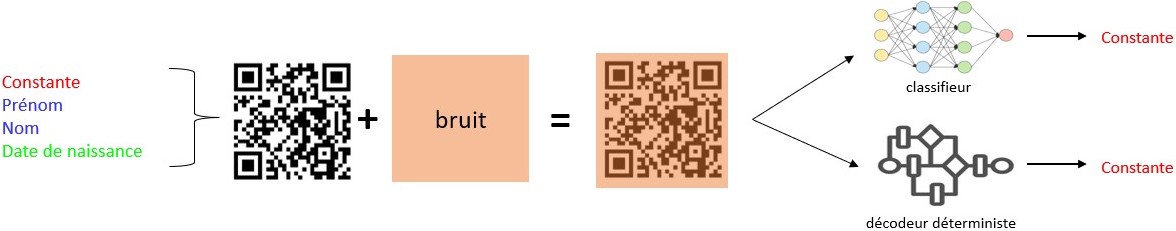}
    \caption{Schéma des expérimentations. 1) Génération des QR codes contenant les informations obtenues lors du pass sanitaire européen. 2) Ajout de bruit. 3) Entraînement d'un classifieur sur les images bruitées. 4) Comparaison des performances du classifieur et celles d'un décodeur classique de QR Codes pour la tâche de détection de la constante de validité du pass.}
    \label{fig:abstract}
\end{figure*}

Il existe des travaux combinant opérateurs classiques de traitement d'images et modèle d'apprentissage basés sur des réseaux de neurones pour prendre en compte les distorsions possibles sur les QR Codes. En effet, la reconnaissance des QR Codes est souvent confrontée à des problèmes de fluctuations irrégulières de l'arrière-plan, d'éclairages inadéquats et de distorsions dues à une méthode d'acquisition d'image difficile. Pour faire face à ces problèmes, des travaux ont vu le jour, que ce soit en adoptant un algorithme de filtre médian adaptatif amélioré et une méthode de correction de distorsion de QR Codes basée sur des réseaux de neurones \cite{Fluctuation}, ou en proposant un algorithme qui localise et segmente les QR Codes à l'aide d'un réseau neuronal convolutif \cite{detecte} notamment pour résoudre d'éventuelles rotations et déformations liées à des scènes complexes \cite{FPN}.\\
Nous souhaitons voir le problème d'un autre point de vue et explorer la possibilité de décoder un QR Code, ou une partie, par le biais d'un modèle d'apprentissage.\\

L'objectif de cette étude est de faire une analyse empirique permettant de définir les limites d'un modèle basé sur l'apprentissage profond pour la classification d'images dites abstraites. En particulier nous posons la question de savoir si un modèle d'apprentissage profond, avec l'introduction d'un a priori sur l'information encodée via les exemples d'apprentissage, peut être plus robuste qu'un modèle déterministe en présence de bruit. Pour cela, nous générons des données de synthèses, des QR Codes contenant les informations obtenues lors du scan d'un pass sanitaire européen soit le prénom, le nom, la date de naissance et une constante relative à la validité du pass (valide ou invalide). Dans ce travail, nous cherchons à prédire cette constante. Nous traitons donc un problème de classification binaire. Enfin, nous ajoutons différents types de bruits aux QR Codes pour comparer la robustesse d'un modèle [neuronal] de classification d'images et celle d'un décodeur de QR Code classique déterministe. Les différentes étapes de l'expérimentation sont schématisées dans la Figure~\ref{fig:abstract}. La méthode de génération des images, l'ajout des différents bruits aux images, la distribution des jeux de données d'entraînement, de validation et de test ainsi que l'explication des méthodes de classification utilisées sont détaillés dans le section \ref{Méthode}. La section \ref{Etude expérimentale} présente les résultats des expérimentations. Enfin, ces résultats sont discutés dans la section \ref{Discussions} dans laquelle nous présentons les questions soulevées par les résultats des expérimentations ainsi que les extensions possibles de ces travaux.

\section{Méthode}
\label{Méthode}
\subsection{Génération des QR Codes}
\begin{figure}
    \centering
    \includegraphics[scale=0.4]{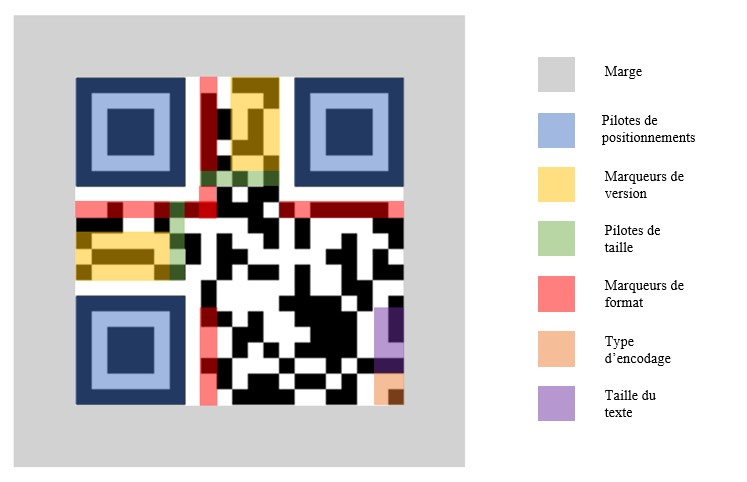}
    \caption{Structure d'un QR Code} 
    \label{fig:structure}
\end{figure}
Les QR Codes sont utilisés pour représenter des informations en les encodant. La structure d'un QR (cf Fig.~\ref{fig:structure}) suit un schéma précis pour faciliter son décodage. Il existe plusieurs versions de QR Codes, en fonction du nombre de caractères encodés.

\begin{figure*}[!ht]
    \centering
    \includegraphics[width=.8\textwidth]{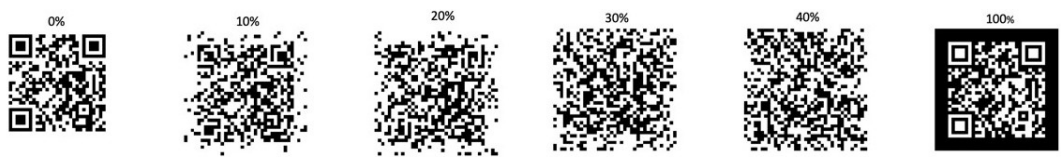}
    \caption{QR codes dont un pourcentage de la surface totale est bruitée en inversant la valeur des pixels (un pixel noir devient blanc et inversement).} 
    \label{fig:inverse}
\end{figure*}

Afin d'assurer la bonne transmission de l'information, les QR Codes sont générés avec un système de correction d'erreur appelé système Reed-Solomon \cite{reed-salomon}. Ce système permet de corriger les erreurs dues à la transmission imparfaite d'un message en répliquant l'information un certain nombre de fois à une certaine distance. Il existe 4 niveaux de correction d'erreur associés à un pourcentage théorique de la surface pouvant être bruitée sans influencer les performances du décodeur. Ces niveaux sont : L (7\%), M (15\%), Q (25\%) et H (30\%).\\

Pour nos expérimentations, nous générons des QR Codes à partir de chaînes de caractères de la forme :\\
\textcolor{red}{constante}/\textcolor{blue}{Prénom}/\textcolor{blue}{Nom}/\textcolor{green}{date\_de\_naissance} avec : \\
\\
\textcolor{red}{constante} $\in \{\textrm{approuvé};\ \textrm{invalide}\}$ \\
\textcolor{blue}{Prénom} et \textcolor{blue}{Nom} tirés aléatoirement avec la bibliothèque \textit{names}\footnote{https://pypi.org/project/names/}\\
\textcolor{green}{date\_de\_naissance} $\in $ [01/01/1921 ; 01/01/2021]\\
\\
Les QR codes sont générés avec la bibliothèque \textit{qrcode}\footnote{https://pypi.org/project/qrcode/} en prenant en paramètres d'entrées la chaîne de caractères et le niveau de correction d'erreur.
\subsection{Modèles de bruit}
Lors de la génération d'un QR Code, si celui-ci a une correction d'erreur de niveau Q alors cela implique que jusqu'à 25\% de sa surface totale peut être bruitée ou cachée sans pour autant empêcher un décodeur classique de le lire correctement. Afin de vérifier cette hypothèse, nous bruitons un certain pourcentage des QR Codes en inversant la valeur des pixels (blanc$\rightarrow$noir, noir$\rightarrow$blanc) (cf Fig.~\ref{fig:inverse}). Pour faciliter l'analyse des résultats, comme le décodeur déterministe lit un QR code bit par bit, nous générons les QR Codes tel que 1 bit corresponde à 1 pixel.\\

\begin{figure}
    \centering
    \includegraphics[width=.9\columnwidth]{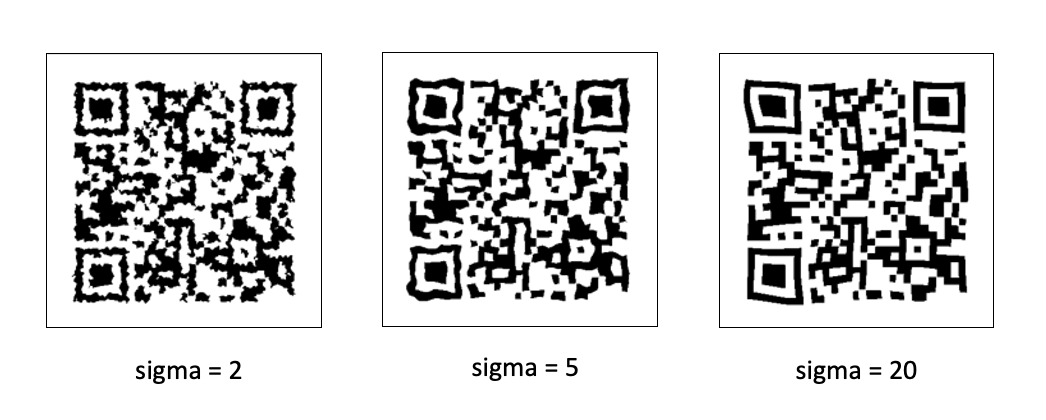}
    \caption{Exemples de QR Codes bruités avec la technique \textit{Random distortions} qui simule l'étalement de l'encre en fonction du paramètre \textit{sigma}.}
    \label{fig:random}
\end{figure}
\begin{figure}
    \centering
    \includegraphics[width=.9\columnwidth]{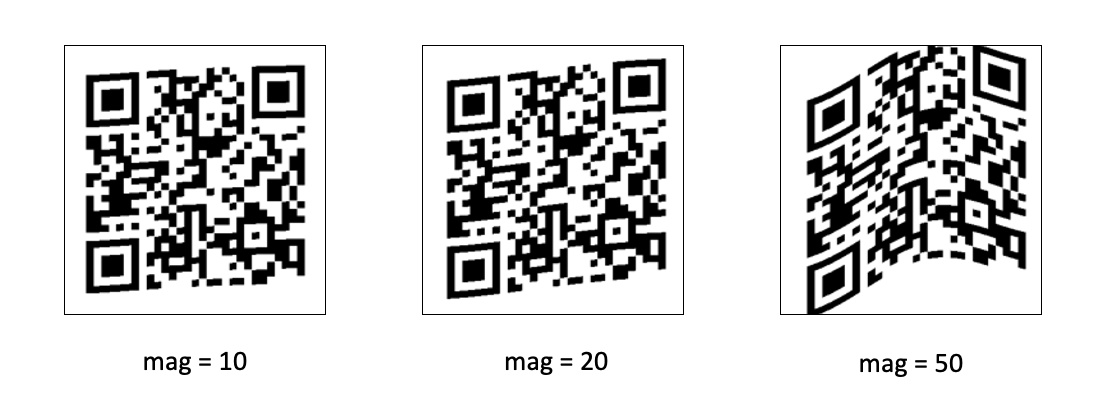}
    \caption{Exemples de QR Codes bruités avec la technique \textit{Ruled surface distortions} en fonction du paramètre \textit{mag}. Cette technique, développée par NVIDIA simule le gondolage du papier.}
    \label{fig:ruled}
\end{figure}
\begin{figure}
    \centering
    \includegraphics[width=.5\columnwidth]{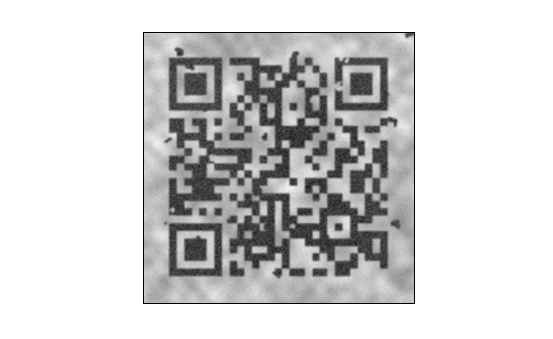}
    \caption{QR Code bruité avec la technique \textit{Foreground/Background selection} de NVIDIA, qui simule une baisse de luminosité.}
    \label{fig:selection}
\end{figure}
L'utilisation d'un décodeur classique étant courante au quotidien, nous souhaitions aussi en évaluer les performances sur des QR Codes bruités de façon plus réaliste que par inversement de pixels. Inspirés par des travaux sur la dégradation de document \cite{kanungo}, nous utilisons \textit{ocrodeg}\footnote{https://github.com/NVlabs/ocrodeg}, une bibliothèque développée par \textit{NVIDIA} permettant de simuler différents bruits sur des documents numérisés \cite{ocrodeg}. Plus précisément, nous utilisons trois types de bruits :\\
\begin{itemize}
    \item \textit{Random distortions} (cf Fig.~\ref{fig:random}), qui simule l'étalement de l'encre en fonction d'un paramètre \textit{sigma}.
    \item \textit{Ruled surface distortions} (cf Fig.~\ref{fig:ruled}), qui simule le gondolage du papier en fonction d'un paramètre \textit{mag}.
    \item \textit{Foreground / Background selection} (cf Fig.~\ref{fig:selection}), qui simule des défauts de luminosité.
\end{itemize}
Lors de l'utilisation du décodeur classique sur les images bruitées avec l'approche \textit{Foreground / Background selection}, le décodage est précédé par une étape de binarisation dont le seuil est déterminé par la méthode Otsu \cite{otsu}. En effet, comme cette approche modifie l'intensité lumineuse des pixels en nuances de gris, le décodeur ne peut pas les décoder sans étape de binarisation préalable.

\subsection{Jeux de données}
\begin{figure}
    \centering
    \includegraphics[scale=0.3]{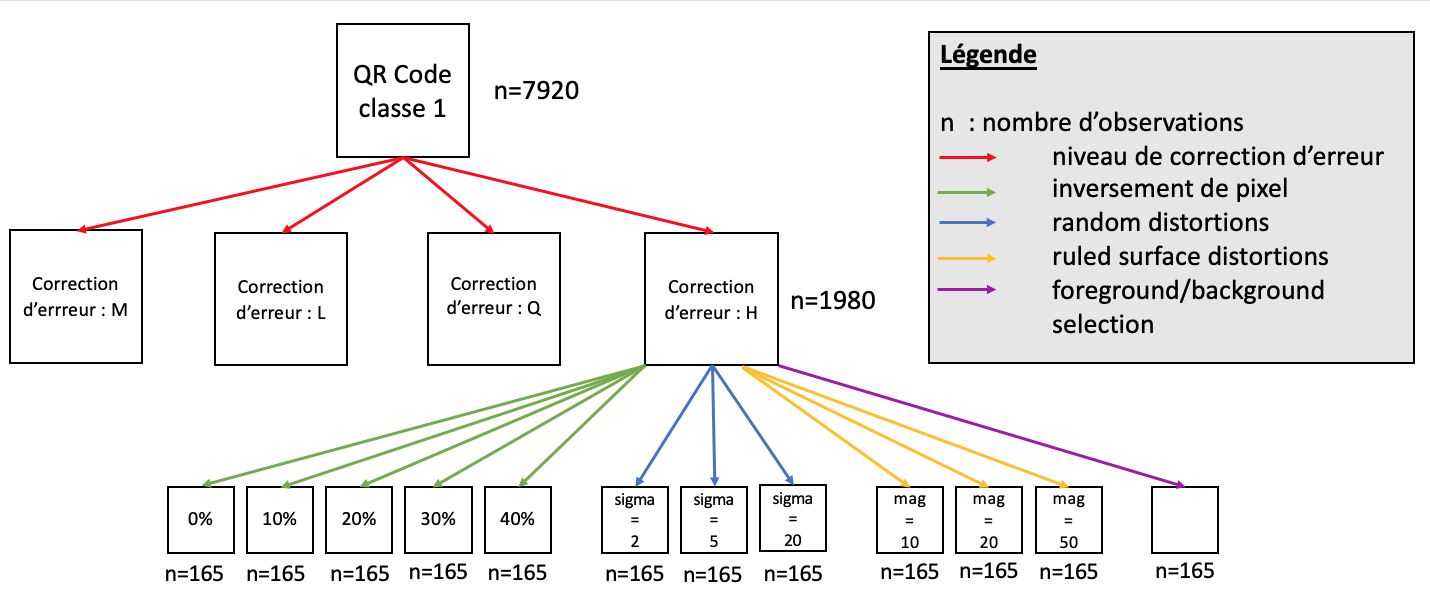}
    \caption{Arbre représentatif de la distribution du jeu de données d'entraînement. Chaque classe est représentée par 7920 observations dont 1980 par correction d'erreur. Pour chaque correction d'erreur, 165 observations sont générées par type de bruit : inversement de pixels de 0\% à 40\% inclus, \textit{Random distortions} pour sigma $\in \{2;5;20\}$, \textit{Ruled surface distortions} pour mag $\in \{10;20;50\}$ et \textit{Foreground/Background selection}}
    \label{fig:distribution}
\end{figure}

Nous entraînons un modèle de classification sur des QR Codes auxquels sont rajoutés les différents modèles de bruit présentés précédemment. Le jeu de données d'entraînement contient 7920 observations par classe (approuvé, invalide), soit 15 840 observations au total. La distribution du jeu de données d'entraînement, dont 20\% est utilisé pour la validation, est décrite dans la Figure ~\ref{fig:distribution}.\\
Pour les tests, nous générons plusieurs jeux de données de 2000 observations chacun (1000 par classe) :\\
\begin{itemize}
    \item jeu de données bruitées par inversement pixels de 0\% à 100\% par paliers de 1.
    \item jeu de données bruitées par la fonction random distortions avec \textit{sigma} $\in \{2;5;20\}$
    \item jeu de données bruitées par la fonction ruled surface distortions avec \textit{mag} $\in \{10; 20; 50\}$
    \item jeu de données bruitées par la fonction foreground / background selection.
\end{itemize}
\subsection{Prédiction de la constante}
L'objectif est de comparer la robustesse d'un modèle de classification d'images basé sur un CNN à celle d'un décodeur déterministe sur des QR Codes bruités.
\paragraph{Classifieur} Nous choisissons d'entraîner un modèle basé sur l'architecture ResNet18 \cite{resnet} sur les QR Codes bruités. Ce choix est motivé par le fait que le modèle ResNet18 est un bon compromis entre performance et taille du réseau.\\
Le modèle est entraîné sur des QR Codes bruités de 0\% à 40\% par paliers de 10\%. Nous faisons le choix de ne pas entraîner le modèle sur des QR Codes bruités au delà de 50\% car il s'agit du seuil de dégradation maximale avant la symétrie. En effet, un QR Code bruité à 100\% est finalement un QR Code négatif sans bruit (cf Fig.~\ref{fig:inverse}).\\
Le modèle est entraîné sur 100 \textit{epochs} pour pouvoir assurer la convergence du modèle et le \textit{batch size} est de 1024, taille maximum pour les ressources GPU que nous avons à disposition. La \textit{loss} utilisée est l'entropie croisée \cite{cross} et le modèle est optimisé à l'aide de l'optimiseur Adam \cite{adam} paramétré par :\\
\begin{itemize}
    \item \textit{learning rate} = 0,0001, qui contrôle le degré d'ajustement des poids du réseau par rapport à la perte de gradient.
    \item \textit{weight decay} = 1e-4, qui ajoute une pénalité à la fonction de perte.
\end{itemize}
De plus, nous réduisons le taux d'apprentissage lorsqu'une métrique a cessé de s'améliorer grâce à la fonction \textit{ReduceLROnPlateau} avec \textit{patience} = 5, qui correspond au nombre d'époques sans amélioration après lesquelles le taux d'apprentissage est réduit.\\

\paragraph{Décodeur} Concernant le décodeur déterministe de QR Codes, nous utilisons un décodeur couramment utilisé \cite{pyzbar1,pyzbar2} de la bibliothèque \textit{pyzbar}\footnote{https://pypi.org/project/pyzbar/} qui implémente les étapes classiques de décodage de QR Codes.
\section{Etude expérimentale}
\label{Etude expérimentale}
\subsection{Méthode d'évaluation}
Nous évaluons, avec la mesure de l'\textit{accuracy}, les performances du classifieur en comparaison avec celles du décodeur classique de QR Codes. L'\textit{accuracy} est le ratio entre le nombre d'observations bien classées par le modèle et le nombre total d'observations à classer. Pour le décodeur, nous estimons qu'il détecte bien si la constante ("invalide" ou "approuv") est dans la chaîne de caractères du résultat\footnote{Pour le cas des QR Codes générés avec la constante "approuvé", dans un souci de correspondance ASCII, nous nous contentons de la séquence "approuv".}.
\subsection{Comparaison des performances des modèles sur des QR Codes bruités aléatoirement}
\begin{figure}
    \centering
    \includegraphics[scale=0.3]{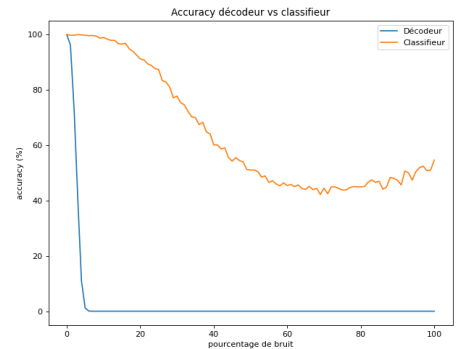}
    \caption{Courbes d'\textit{accuracy} du décodeur déterministe et du classifieur sur des QR Codes bruités de 0\% à 100\%}
    \label{fig:bruit}
\end{figure}
La Figure~\ref{fig:bruit} montre les comportements du classifieur ainsi que celles du décodeur sur des images bruitées par inversement de pixels de 0\% à 100\%. On remarque qu'à partir de 50\% de bruit (correspondant à un QR code aléatoire), l'\textit{accuracy} du classifeur est proche de 50\% ce qui correspond effectivement à de l'aléatoire. Concernant le décodeur, ses performances sont nulles à partir de 6\% de bruit. Ce résultat est dû au fait que le bruit ajouté peut se trouver dans les pilotes de positionnements. Les pilotes de positionnements sont les grands carrés aux trois coins (cf Fig.~\ref{fig:structure}) du QR Code qui déterminent sa position et son orientation. Pour décoder un QR Code, la première étape faite par le décodeur consiste à détecter ces pilotes de positionnements. Donc si le décodeur n'arrive pas à les identifier, il ne peut pas considérer l'image comme un QR Code et le décoder, ce qui n'est pas le cas du classifieur. Il est à noter que des expérimentations \cite{markeurs} ont montré que si les pilotes de positionnements ne sont pas bruités alors la surface maximale pouvant être bruitée correspond bien au niveau de correction d'erreur.
\begin{table}
\caption{{Comparaison des performances du classifieur et du décodeur en fonction du bruit utilisé durant la génération des QR Codes. Les niveaux de bruit, paramétrés par différentes valeurs, sont moyennés.}}
\begin{center}
\begin{tabular}{|c|c|c|c|}
 
 \hline
     Bruit & Classifieur & Décodeur  \\
     \hline
     \footnotesize{Random distortions} & 99,5\% & 63,5\% \\
     \hline
    \footnotesize{Ruled surface distortions} & 99,35\% & 54,3\% \\
     \hline
     \footnotesize{Foreground/Background selection} & 99,35\% & 97,5\%\\
     \hline
\end{tabular}
\label{tab1}
\end{center}
\end{table}
\subsection{Comparaison des performances des modèles sur des QR Codes en fonction du bruit réaliste appliqué}
La Table \ref{tab1} présente les performances des deux méthodes en fonction du bruit utilisé. Ces bruits sont plus réalistes car correspondant à des défauts d'impression ou de capture de l'image. Nous observons que le classifieur est très performant quel que soit le bruit. En comparaison, le décodeur arrive à détecter presque deux tiers des QR Codes bruités avec \textit{Random distortions} et seulement la moitié des QR Codes bruités avec \textit{Ruled surface distortions}. Enfin, 97,5\% des QR Codes bruités avec \textit{Foreground/Background selection} sont détectés par le décodeur. Bien que pour toutes les expérimentations effectuées, le CNN détecte bien à minima tous les QR Codes bien détectés par le décodeur déterministe, il est important de noter que ces résultats ne reflètent pas nécessairement ceux que pourraient avoir les décodeurs utilisés au quotidien, notamment ceux inclus dans les smartphones. Ces derniers sont généralement couplés à des techniques de pré-traitement permettant une meilleure lecture du QR Code.
\subsection{Tests sur des QR Codes encodant d'autres informations}
\begin{figure}
    \centering
    \includegraphics[scale=0.3]{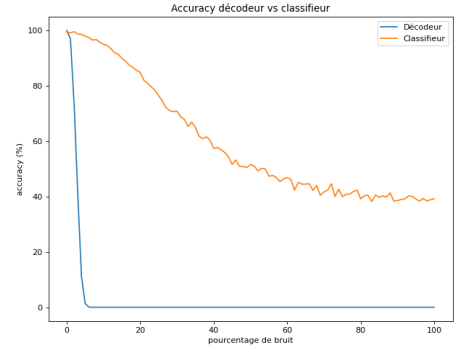}
    \caption{Courbes représentatives de l'\textit{accuracy} du décodeur déterministe et du classifieur sur des QR Codes générés à partir de séquence de type https://constante/username avec constante = Facebook ou Twitter.}
    \label{fig:fb/twit}
\end{figure}

Finalement, pour pouvoir généraliser nos expérimentations, nous avons effectué les mêmes tests en encodant des chaînes de caractères correspondant à des URls vers des comptes Facebook ou Twitter de la forme : https://\textcolor{red}{constante}/\textcolor{blue}{Username} avec :\\
\textcolor{red}{constante} = Facebook ou Twitter\\
\textcolor{blue}{Username} chaîne de $x\in[5;20]$ caractères tirés aléatoirement.\\
La Figure~\ref{fig:fb/twit} montre que les résultats sont comparables à ceux présentés auparavant et confirme la possibilité de généraliser notre approche.\\
\section{Discussion}
\label{Discussions}
\subsection{Similarité syntaxique et positionnement}
Les informations obtenues lors du scan d'un pass sanitaire sont le nom, le prénom, la date de naissance et une constante validant ou non le pass. Nous avons généré des QR Codes à partir de chaînes de caractères contenant ces informations. Nous avons testé différentes constantes pour analyser l'influence de la similarité syntaxique entre les mots. Les constantes de classes utilisées sont :
\begin{itemize}
    \item valide / invalide 
    \item true / false
    \item approuvé / invalide
\end{itemize}
Pour chaque constante, le format de la chaîne de caractères est : \textcolor{blue}{Prénom}/\textcolor{blue}{Nom}/\textcolor{green}{date\_de\_naissance}/\textcolor{red}{constante}.\\
La génération d'un QR Code est une suite d'étapes prédéfinies et n'inclut pas la notion d'aléatoire (le résultat de génération de QR codes à partir de deux séquences identiques est le même). Ainsi, si un caractère est toujours au même endroit dans la séquence, l'octet correspondant sera toujours au même endroit dans le QR Code. La place de la constante à prédire dans la chaîne encodée a donc une influence forte sur les performances d'un modèle basé CNN. Les variables nom et prénom n'étant de longueur fixe, la constante ne se trouve pas toujours au même endroit. Nous avons donc généré, pour la constante ayant donné de meilleurs résultats, une séquence de la forme : \textcolor{red}{constante}/\textcolor{blue}{Prénom}/\textcolor{blue}{Nom}/\textcolor{green}{date\_de\_naissance}.\\
Pour ces expérimentations, nous avons généré des QR Codes à partir de chaînes de caractères de différentes configurations et avons ajouté du bruit en inversant 0\% à 40\% des pixels, par paliers de 10. Nous testons ensuite le modèle sur des QR Codes bruités de 0\% à 100\% par paliers de 1.\\
\begin{figure}
    \centering
    \includegraphics[scale=0.5]{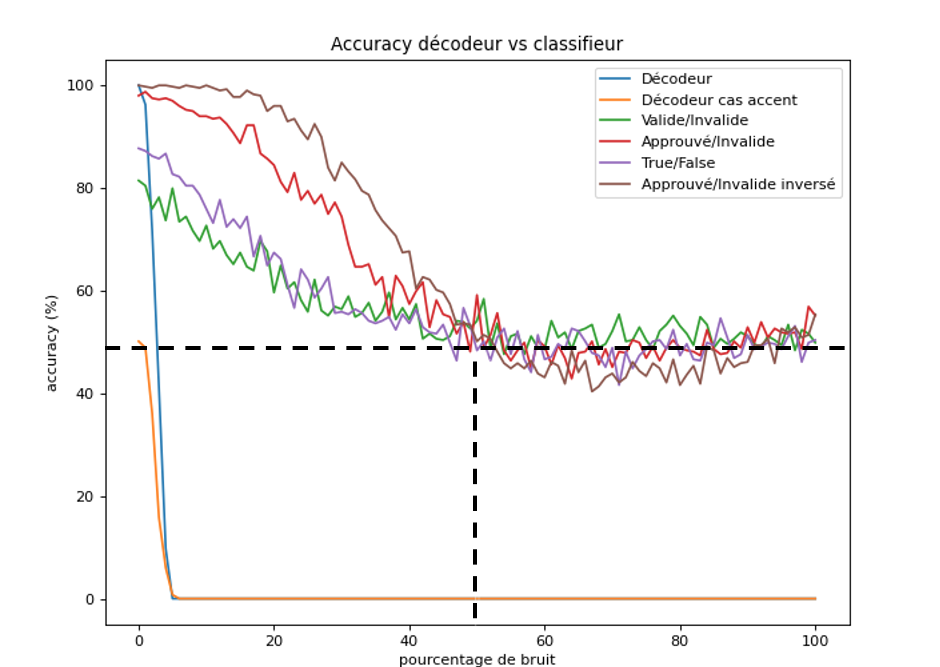}
    \caption{Comparaison de l'inférence des modèles entraînés sur différentes classes en fonction du pourcentage de bruit par inversement de pixels (de 0\% à 100\%).}
    \label{fig:inf}
\end{figure}
\\La Figure~\ref{fig:inf} montre les performances des différents modèles de classification d'images ainsi que les performances du décodeur classique. La courbe bleu correspond aux performances du décodeur déterministe lorsqu'on lui demande de trouver "approuv". Celle orange représente le cas où le décodeur doit trouver "approuvé", ce qu'il n'arrive pas à faire à cause de la correspondance ASCII. Les courbes verte, rouge et violette correspondent respectivement aux performances des classifeurs entraînés à détecter les classes "valide"/"invalide", "approuvé"/"invalide" et "true"/"false" lorsque la classe est placée en fin de séquence pour la génération du QR Code. Enfin, la courbe marron représente les performances du classifieur entraîné à détecter les constantes "approuvé"/"invalide" lorsque qu'elles sont placées en début de séquence pour l'encodage vers un QR Code.\\
Nous observons par ce graphique que la similarité syntaxique entre les mots influe sur les performances du modèle. En effet, pour les séquences ayant le même format, celle qui amène les meilleurs résultats (courbe rouge) contient les mots "approuvé" ou "invalide", qui sont plus ou moins similaires syntaxiquement que les autres. Enfin, pour les constantes "approuvé" et "invalide", les résultats montrent que le modèle est plus performant lorsqu'il est entraîné sur des séquences contenant la constante au début, à une position fixe.\\
Les expérimentations effectuées montrent qu'un modèle couramment utilisé de classification d'images peut être plus performant qu'un modèle déterministe dans le cas contraint de la détection d'une constante. Nos travaux ne constituent pas l'état de l'art pour la tâche de décodage des QR Codes car le modèle de classification d'images n'a pas été entraîné pour détecter la totalité de la chaîne de caractères encodée mais pour localiser et prédire une information jugée importante dans le QR Code. Cependant, ces expérimentations montrent qu'un modèle simple de classification présente de bonnes performances sur des images dîtes "abstraites" et bruitées.\\
\subsection{Extension}
\begin{figure}
    \centering
    \includegraphics[scale=0.42]{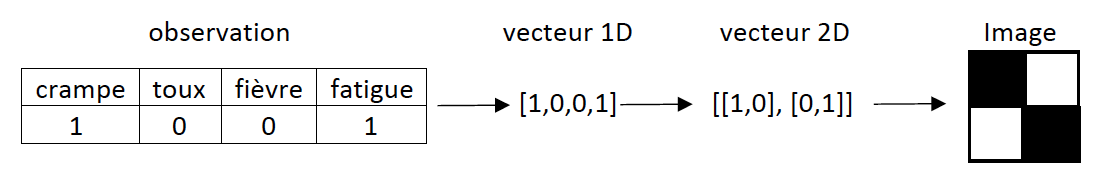}
    \caption{Processus de génération d'images à partir d'une observation.}
    \label{fig:process}
\end{figure}

\begin{figure}
\centering
\includegraphics[scale=0.26]{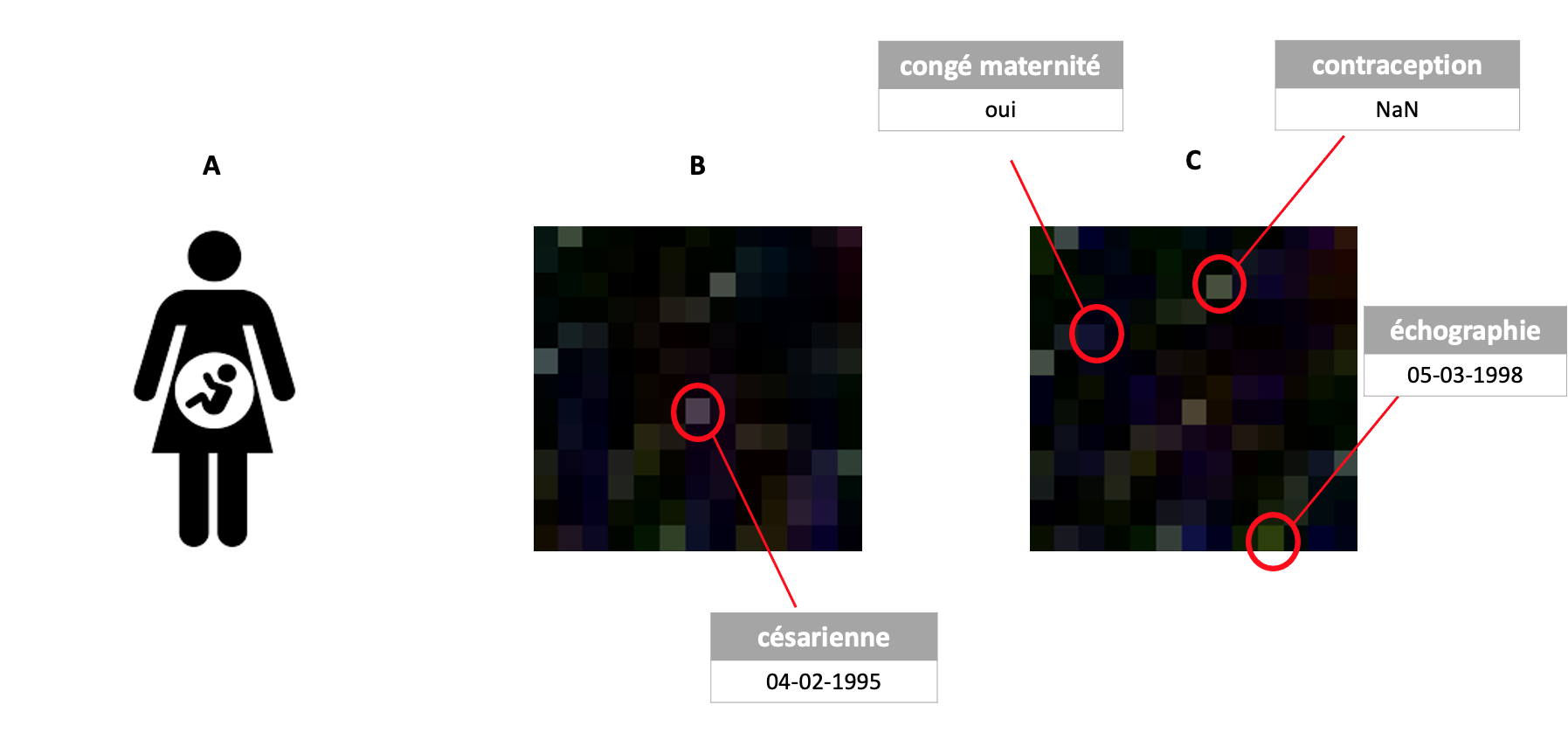}
\caption{Trois images qui correspondent à la classe "a été enceinte". L'image A représente une personne par le dessin de sa silhouette. Les images B et C représentent une grossesse par des pixels correspondant à des événements liés au suivi de la grossesse.}
\label{enceinte}
\end{figure}

L'utilisation de modèles du domaine de la vision par ordinateur sur des images abstraites peut s'appliquer dans la domaine de recherche lié au fléau de la dimension (\textit{curse of dimensionality}) \cite{curse}. Prenons les données du Système National des Données de Santé (SNDS)\footnote{https://www.snds.gouv.fr/SNDS/Accueil} qui regroupent les données de l'assurance maladie, les données des hôpitaux et les causes de décès des personnes résidants en France. Le SNDS est représenté par 200 tables et plus de 3 000 variables descriptives. Dans la plupart des tables, une observation correspond à un évènement dans le parcours médical d'un individu. Autrement dit, un individu est représenté par autant d'observations que d'évènements médicaux constituant son parcours médical. Si on souhaite faire une classification sur les données du SNDS, pour détecter une pathologie par exemple, nous faisons face au phénomène du fléau de la dimension. Les méthodes les plus couramment utilisées pour pallier le fléau de la dimension consistent à réduire les dimensions du jeu de données \cite{dimension}. Ces méthodes permettent d'entraîner les modèles classiques de classification en temps raisonnable mais engendre une perte d'information. Nous souhaitons rester exhaustif dans l'exploitation des données du SNDS et ne pas avoir recours à l'extraction de variables d'importances avant l'entraînement d'un modèle.\\
Inspirés par des travaux sur la génération d'images à partir de données originellement tabulaires, qui utilisent des métriques sur les données telles que les diagrammes à barres équidistantes et la matrice de distance normalisée pour diagnostiquer le cancer du sein \cite{cancer} ou transforment des fichiers binaires en images pour la détection de logiciels malveillants \cite{malware}, nous souhaitons développer le même type d'approche et l'appliquer aux données du SNDS. La transposition des tableaux de données en images permettrait de contourner les approches classiques de réduction de dimension et de ne pas fixer de limite à la taille du jeu de données d'apprentissage (cf Fig~\ref{fig:process}).\\
L'idée est donc de représenter les données du SNDS sous forme d'images pour y appliquer des modèles du domaine de la vision par ordinateur. Il s'agit d'associer à chaque individu une image retraçant son parcours médical. Cette image est une grille pixellique dont la position et la couleur des pixels correspondent à un type d'évènement médical défini au préalable (cf Fig.~\ref{enceinte}).\\
\section{Conclusion}
Les travaux présentés montrent qu'un modèle de classification d'images basé sur l'apprentissage profond, ResNet18, est très performant sur des images dites "abstraites" dont le format est constant et auxquelles du bruit de différentes natures a été ajouté. Ces résultats sont encourageants quant à l'utilisation de modèle du domaine de la vision par ordinateur sur des images dans lesquelles chaque pixel encode une information. Aussi, nous avons montré que ces modèles imposent un certain nombres de contraintes sur l'image abstraite. En particulier, la localisation de l'information doit être constante.\\
Par la suite, nous allons générer les images pour chaque individu présent dans les données du SNDS. L'objectif est de trouver la méthode de génération permettant de rester exhaustif dans la temporalité des évènements tout en maximisant les performances du modèle utilisé.

\end{document}